\documentclass[conference]{IEEEtran}
\IEEEoverridecommandlockouts
\usepackage{cite}
\usepackage{subfigure}
\usepackage{amsmath,amssymb,amsfonts}
\usepackage{algorithmic}
\usepackage{graphicx}
\usepackage{textcomp}
\usepackage{xcolor}
\usepackage{multirow}
\def\BibTeX{{\rm B\kern-.05em{\sc i\kern-.025em b}\kern-.08em
    T\kern-.1667em\lower.7ex\hbox{E}\kern-.125emX}}

\makeatletter

\let\old@ps@IEEEtitlepagestyle\ps@IEEEtitlepagestyle
\def\confheader#1{%
    \def\ps@IEEEtitlepagestyle{%
        \old@ps@IEEEtitlepagestyle%
        \def\@oddhead{\strut\hfill#1\hfill\strut}%
        \def\@evenhead{\strut\hfill#1\hfill\strut}%
    }%
    \ps@headings%
}
\makeatother

 \usepackage[pscoord]{eso-pic}
\newcommand{\placetextbox}[3]{
 \setbox0=\hbox{#3}
 \AddToShipoutPictureFG*{ \put(\LenToUnit{#1\paperwidth},\LenToUnit{#2\paperheight}){\vtop{{\null}\makebox[0pt][c]{#3}}}
 }
 }
 \placetextbox{.23}{0.055}{\small{978-1-6654-2363-2/21/\$31.00 ©2021 IEEE}}

\begin{document}

\title{Fine-Grained Image Generation from Bangla Text Description using Attentional Generative Adversarial Network
}

\author{\IEEEauthorblockN{Md Aminul Haque Palash\textsuperscript{1}, Md Abdullah Al Nasim\textsuperscript{2}, Aditi Dhali\textsuperscript{3} and Faria Afrin\textsuperscript{4}}
\IEEEauthorblockA{\textit{Department of Research and Development; }
\textit{Pioneer Alpha\textsuperscript{1,2,3,4}}
; Dhaka, Bangladesh \\
u1404103@student.cuet.ac.bd\textsuperscript{1},
nasim.abdullah@ieee.org\textsuperscript{2}, aditi.stu2015@juniv.edu\textsuperscript{3} and
faria15-9230@diu.edu.bd\textsuperscript{4}}
}

\maketitle

\begin{abstract}
Generating fine-grained, realistic images from text has many applications in the visual and semantic realm. Considering that, we propose Bangla Attentional Generative Adversarial Network (AttnGAN) that allows intensified, multi-stage processing for high-resolution Bangla text-to-image generation. Our model can integrate the most specific details at different sub-regions of the image. We distinctively concentrate on the relevant words in the natural language description. This framework has achieved a better inception score of 3.58 ± .06 on the CUB dataset. For the first time, a fine-grained image is generated from Bangla text using attentional GAN. Bangla has achieved 7th position among 100 most spoken languages. This inspires us to explicitly focus on this language, which will ensure the inevitable need of many people. Moreover, Bangla has a more complex syntactic structure and less natural language processing resource that validates our work more.

%The proposed AttnGAN significantly outperforms the previous state-of-the-art, boosting the best-reported inception score by 14.14% on the CUB dataset and 170.25% on the more challenging COCO dataset. A detailed analysis is also performed by visualizing the attention layers of the AttnGAN. It, for the first time, shows that the layered attentional GAN is able to automatically select the condition at the word level for generating different parts of the image.

\end{abstract}


\section{Introduction}

In recent years, image processing has become a promising sector to explore. Its vast application area has encouraged more and more researchers every day working in this field. Nowadays, its application is not limited to only processing daily camera-captured images. Rather, image processing is leading a  very important role in the medical field, agricultural sector, and even in educational areas. Combining Natural image processing and digital image processing, a new thriving sector is emerging, which impacts highly in the research areas. In this paper, we work on the "Bangla" natural language and generate high-resolution images (text-to-image converter) using attention GAN, a useful element of image processing. 

The application of Generative Adversarial Networks (GANs) \cite{goodfellow2014generative} has increased dramatically in recent years that can generate well-constructed information at different sub parts of the captured image using a unique attentional generative network. In the deep learning filed, it has gained a lot of attention for the variety of its application and its recent popularity such as images \cite{isola2017image},\cite{zhu2017unpaired},\cite{yi2017dualgan}, texts \cite{ reed2016generative},\cite{ zhang2017stackgan},
\cite{ xu2018attngan},\cite{ zhang2018photographic} and generating images based on discrete labels \cite{ miyato2018cgans},\cite{ odena2017conditional}.

\begin{figure}[h]
\centering
\includegraphics[height=200pt, width=255pt]{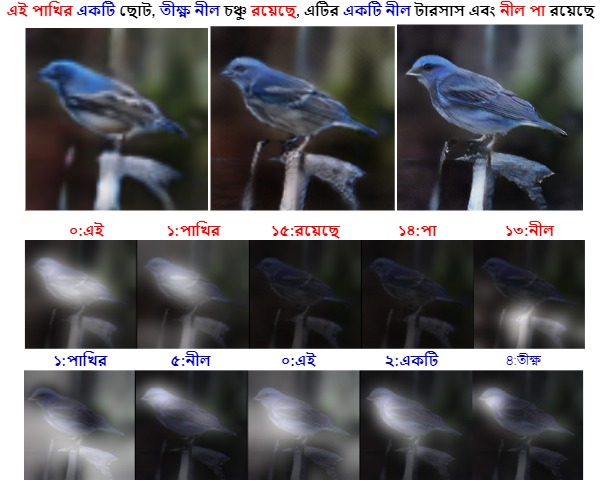}
\caption{Experiment with our proposed model's outcomes. The first line displays low-to high-resolution images, while the second and third lines display the top five most-visited terms..}
\label{f1}
\end{figure}

Generative Adversarial Networks (GANs) \cite{goodfellow2014generative} are the latest text-to-image generation algorithms to be suggested and employed to produce life-like images based on text depiction (see Fig. \ref{f1}). It has introduced a new mechanism for capturing structured word level and multi-staged text-to-image synthesis; we propose Bangla Attentional Generative Adversarial Network (AttnGAN) for synthesizing intricate scenes.
Multi-stage approaches \cite{ zhang2017stackgan}, \cite{ xu2018attngan}, \cite{ zhang2018stackgan++} provide low-resolution beginning images, which are then refined to high-resolution images.

% Fig.1 Example results of the proposed AttnGAN

In our paper, we propose a Bangla attentional GAN for well-constructed text-to-image creation that supports attention-driven and several-stage processing inspired by this work \cite{ xu2018attngan}.
Experiments were performed to synthesize images from text descriptions, including the two most-used components AttnGAN and DAMSM, by translating the CUB \cite{ wah2011caltech}  dataset into Bangla or Bengali. To experimentally evaluate the proposed Bangla AttnGAN, a comprehensive investigation is carried out, which outperforms the previous state-of-the-art model and enhances the quality of generated images. 
Our contributions are as follows:
\begin{itemize}
    \item We generate a text-to-image converter which generates high-definition (256x256), fine-grained image
    \item We are the first to work on Bangla language text-to-image generation

\end{itemize}

% \begin{figure}[h]
% \centering
% \includegraphics[width=0.1\textwidth]{images/i1.jpg}
% \caption{License plate of a vehicle}
% \label{f1}
% \end{figure}

\section{Related Works}
Creating images analyzing from natural language is one of the essential uses of ongoing conditional generative models.
In art generation, image editing (brightening, enhancing), computer-aided design, and many others are the amazing applications of generating well-constructed, life-like, high-resolution images from text descriptions. Recently, incredible advancement has been accomplished toward this path with the rise of deep generative models. A combination of deep architecture and deep convolutional generative adversarial networks (GANs) formulation can effectively bridge the advancement in text and image modeling as GAN generate highly compelling images \cite{reed2016generative}. MirrorGAN considered both global and local attention, and they preserve semantic coherence to generate text-to-image \cite{qiao2019mirrorgan}. A semantic text embedding module (STEM), a global-local collaborative attentive module (GLAM), and a semantic text regeneration and alignment module (STREAM) are used to build this model. AttnGAN is another proposed model to generate fine-grained text-to-image \cite{xu2018attngan}. This model then is improved by Naveen et al. by solving the extraction of semantic information from the text descriptions combining BERT, GPT2, XLNet with AttnGAN \cite{naveen2021transformer}. Additionally, it permits attention-driven, multi-stage refinement for this type of image generation. Sharma et al. \cite{sharma2018chatpainter} came up with a new idea of adding dialogue after text-to-image generation for significant progress in the inception score. ControlGAN is word-level spatial and channel-wise attention-driven text-to-image generator \cite{li2019controllable}. Yin et al. \cite{yin2019semantics} proposed another photo-realistic text-to-image generator that fulfills both the high-level and low-level semantic consistency. Attention loss and diversity loss are used to enhance the sensitivity of the GAN \cite{hu2021crd}. 

On the other hand, the Zero-Shot framework models the text and image tokens as a single stream of data \cite{ramesh2021zero}. In the ManiGAN framework, an image is semantically edited part by part matching a given text focusing on the desired attributes \cite{li2020manigan}. TediGAN was proposed by Xia et al. for multi-modal image generation. This model can effectively manipulate textual descriptions \cite{xia2021tedigan}. Another new model of generating face images from text is modeled by khan et al. \cite{khan2020realistic}. Schulze et al. generate text-to-image using combined GAN. It can initiate photo-realistic images described as textual descriptions \cite{schulze2021cagan}.

\section{Methodology}

%\subsection{Generative Adversarial Network (GAN)}

%A Generative Adversarial Network, or GAN, is a sort of neural network architecture for generative modeling. 
%Generative modeling includes utilizing a model to produce new models that conceivably come from an existing distribution of samples, for example, creating new photos that are comparative however explicitly unique in relation to a dataset of existing photos. 
GAN is a generative model that is trained consisting of two neural network models named "generator" or "generative network" and "discriminator" or "discriminative network" model in most cases. The "generative network" model successfully measures how to generate fresh conceivable samples. The "discriminative network" model, on the other hand, specifies how to distinguish between fabricated and actual cases.

The attentional generative network and the deep attentional multi-modal similarity model are two distinct components of our proposed Bangla Attentional Generative Adversarial Network (AttnGAN).

%This model gives an approach to learn deep representations without broadly annotated training data. They accomplish this by inferring backpropagation signals through a competitive cycle, including a couple of networks. The representations that can be learned by GANs might be utilized in image synthesis, image analysis, semantic image altering, classification, image superresolution, and style transfer \cite{8253599}. GAN has been modified later according to researchers' needs and to output various classified images. Unrolled Generative Adversarial Networks \cite{metz2016unrolled}, Least Squares Generative Adversarial Networks \cline{mao2017least}, Coupled Generative Adversarial Networks \cite{liu2016coupled}, Self-attention generative adversarial networks \cite{zhang2019self} are some of the examples of this modification. Other than that, TAGAN works as a text adaptive network to manipulate the image with natural language \cite{nam2018text}. Zhao et al. proposed a methodology to convert text-to-remote-sensing-image using GAN \cite{zhao2021text}.

\subsection{Attentional Generative Adversarial Networks} 

The Attention Generative Adversarial Network enables attention-driven, long-range dependency modeling for picture generation tasks from Bangla Text. In wide-range resolution image analysis, conventional convolutional GANs provide fine-grained nuances as a component of only spatially local points. Subtleties can be constructed using signals from all attribute locations in the Attention Generative Adversarial Network. Furthermore, the discriminator is used to ensure the consistency of the fine-detailed feature pieces in far-flung portions of the image. The prerequisite for image generation in current GAN-based models for text-to-image generation is to encode the full sentence text portrayal into a single vector, although high-resolution word-level data is required. The generative networks in our model are able to create distinct subregions of the picture based on the phrases that are most relevant to those subregions.
The three generators in our proposed Bangla AttnGAN accept the hidden states as input and generate images on small to big scales as output. The Text Encoder encodes phrase and word features as they flow through the model at various stages. Conditioning Augmentation is used for the conversion between the sentence vector and the conditioning vector.

The word features and the image features from the preceding hidden layer are both inputs to the attention model. By introducing a new perceptron layer, the word features are first translated into the common semantic space of the image data. Then, depending on the image's hidden features, a word-context vector is computed for each sub-region.
At the next stage, with the combination of features extracted from the images and associated text features, images are generated. The attentional generating network's final objective function is defined as to create realistic visuals with several levels (i.e., word level and  sentence level) of circumstances; the attentional generative network's final objective function is defined as:
\begin{equation}
    L=L_G+\beta L{_{DAMSM}}
\end{equation}
where, 
\begin{equation}
   L_G = \sum_{i=0}^{n-1}L_{G_i}
\end{equation}
Here, $beta$ is a hyperparameter that balances the two terms of Eq. (1). The first part represents the GAN loss that approximates conditional and unconditional distributions together. Along with its corresponding discriminator $D_i$, the adversarial loss for $G_i$ is defined with the combination of unconditional loss and conditional loss. Unconditional loss is utilized to identify whether the image is natural or counterfeit, and the conditional loss evaluates the similarity of the image and the sentence.

\begin{figure*}[h]
\centering
\includegraphics[height=200pt, width=500pt]{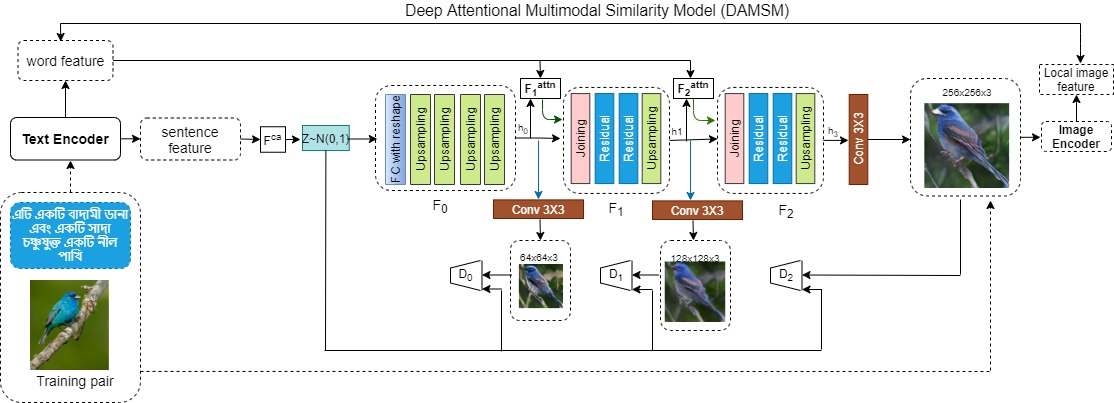}
\caption{Overall architecture of our proposed model}
\label{f2}
\end{figure*}

%This model effectively enhances image processing techniques. Creative fashion generation from natural language \cite{yangcreative} and hybrid attention driven text-to-image are some of them \cite{cheng2019hybrid}. 
Our Bangla Attention Generative Adversarial Network finds the high-resolution output from Bangla language text and sentences. Our overall architecture is shown in Fig. \ref{f2}.

\subsection{Deep Attentional Multimodal Similarity Model}
The DAMSM investigates two neural networks that design sub-regions of natural language text from a sentence and an image to a common semantic space, calculating image-sentence matching compatibility at the word level to determine a well-measured loss for image production.

\subsubsection{Text encoder}
The text encoder is a two-way system Long Short-Term Memory (LSTM) \cite{huang2015bidirectional} that extracts grammatical vectors from the text representation. Hence, we link its two states (hidden) to address the linguistic translation of a word. 

\subsubsection{Image encoder}
Image encoder is Convolutional Neural Network
(CNN) for mapping captured images into the grammatical structure. In this case, images are partitioned into several sub-regions. The intermediary CNN layers work with the local attributes of various sub-parts of the prime image. On the other hand, the last layers explore the global attributes of the image.
\subsubsection{Attention-driven image-text matching score}
The attention-driven image-text matching score aims to quantify the coordination of an image-text pair as a result of a text-to-image attention model. Finally, images for the following step are created by merging image attributes with the relevant text-context data.

\begin{figure*}[h]
\centering
\includegraphics[height=250pt, width=400pt]{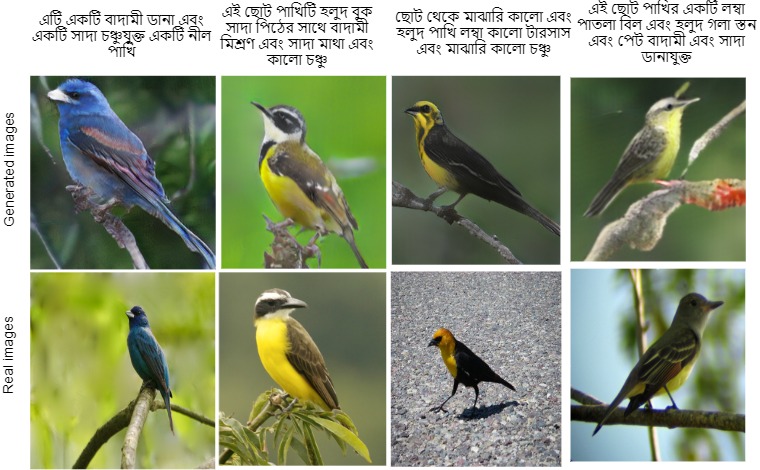}
\caption{Comparison between our model generated images and real images}
\label{syn-real}
\end{figure*}

\begin{figure}[h]
\centering
\includegraphics[height=225pt, width=225pt]{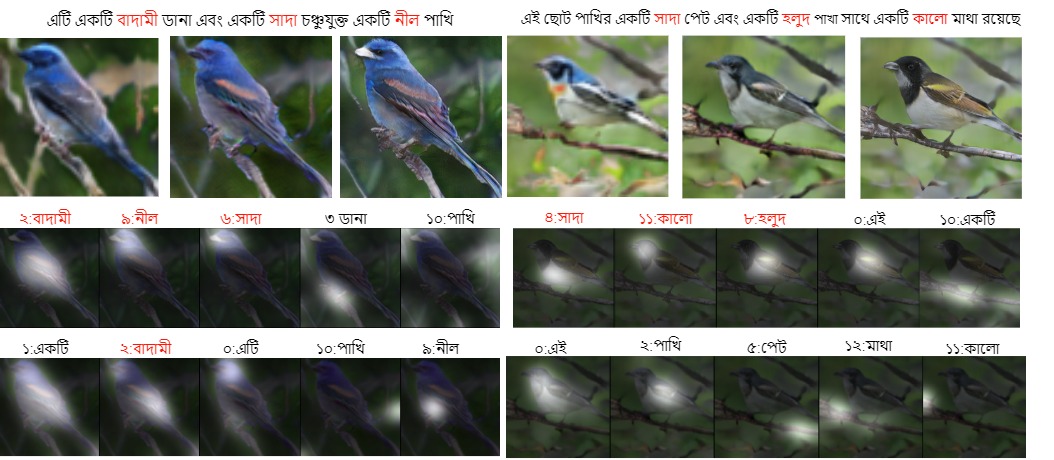}
\caption{More examples of our AttnGAN on CUB as a supplemental material. The first line gives the low-to-high resolution pictures; the second and third line shows the top-5 most-attended words.}
\label{sada-kalo-img}
\end{figure} 

%To produce practical images with multiple dimensions (i.e., word-level and sentence-level) as per as conditions, the final objective function of the attentional generative organization is characterized as:
%\begin{equation}
%    L=L_G+\beta L{_{DAMSM}}
%\end{equation}
%where, 
%%   L_G = \sum_{i=0}^{m-1}L_{G_i}
%\end{equation}
%Here, $beta$ is a hyperparameter that balances the two terms of Eq. (1). The first part represents the GAN loss that approximates conditional and unconditional distributions together. Along with its corresponding discriminator $D_i$, the adversarial loss for $G_i$ is defined with the combination of unconditional loss and conditional loss. Unconditional loss is utilized to identify whether the image is real or fake, and conditional loss evaluates the similarity of the image and the sentence.

Then, we do the loss minimization to group the input into the realm of natural or fake images according to AttnGAN \cite{xu2018attngan}.

\subsubsection{DAMSM Loss}
 When comparing images and text descriptions, we use the DAMSM loss \cite{xu2018attngan}. The DAMSM loss is important because it improves the condition of generated images with text descriptions.
%DAMSM loss is defined as \cite{xu2018attngan}:
%\begin{equation}
%L{_{DAMSM}} = {L_{1}}^{w}+{L_{2}}^{w}+{L_{1}}^{s}+ {L_{2}}^{s}
%\end{equation}
%\begin{equation}
%{L_{2}}^{w}= -\sum_{i=1}^{M} log P\left ( Q_{i} |D_{i}\right  )
%\end{equation}
%\begin{equation}
%P\left ( Q_{i} |D_{i}\right  ) = \frac{exp(\gamma_3R (Q_{i} |D_{i})  ) }{\sum_{j=1}^{M} exp(\gamma_3R (Q_{j} |D_{i})  )}
%\end{equation}

%where, 
%$P\left ( Q_{i} |D_{i}\right  )$ represents the posterior probability of sentence $D_i$ being matching with image $Q_i$ for a batch of image-sentence pairs $\left ( Q_{i} |D_{i}\right  )_{i=1}^{M}$.

%$\gamma_3$ is the smoothing factor

%$w$ represents the word, and $s$ is used to represent sentences. 

\section{Machine configuration}
We do our experiment on a computer with GPU Core 1, CPU Core 4, Ram 61 GB. Details of our used machines are presented in Table \ref{Machine config.} 

\begin{table}[]
\centering
\caption{Machine details}
\label{Machine config.}
\begin{tabular}{|c|c|c|c|c|}
\hline
\textbf{Name}      & \textbf{GPUs} & \textbf{vCPUs} & \textbf{RAM (GiB)} & \textbf{Network Bandwidth} \\ \hline
\textbf{p2.xlarge} & 1             & 4              & 61                 & High                       \\ \hline
\end{tabular}
\end{table}

\section{Experimental Result}
We do extensive quantitative and qualitative evaluations to validate our method. We have done a large-scale experimental analysis effectively to evaluate our proposed model Bangla AttnGAN. Firstly, we explore AttnGAN and the DAMSM. After that, we make a comparison with previous GAN models mentioned earlier for text-to-image analysis. We trained DAMSM model 200 epochs and AttnGAN model 600 epochs.
\subsection{Dataset}
Same as previous text-to-image methods \cite{reed2016learning}, \cite{reed2016generative}, \cite{zhang2017stackgan}, \cite{zhang2018stackgan++}, \cite{xu2018attngan}, CUB \cite{wah2011caltech} is used to evaluate our proposed method. This dataset contains 200 categories bird images and 10 captions per image \cite{reed2016learning}. Table \ref{dataset table} lists the details of this dataset. By using test:train data ration 70:30, we get our best result. In that case, we achieve FID score of 41.08 which is better than the other.

Some other methodology has used Coco dataset \cite{lin2014microsoft}. We are still in the process of modifying this dataset suitable to use in the Bangla language.

% \usepackage{multirow}
\begin{table}[]
\centering
\caption{Statistics of dataset}
\label{dataset table}
\begin{tabular}{|c|c|c|l|l|l|l|}
\hline
\multirow{3}{*}{\textbf{Dataset}} & \multicolumn{6}{c|}{\textbf{CUB dataset}}                                                                                  \\ \cline{2-7} 
                                  & \multicolumn{2}{c|}{Split Ratio: 70:30} & \multicolumn{2}{c|}{Split Ratio: 80:20} & \multicolumn{2}{l|}{Split Ratio:60:40} \\ \cline{2-7} 
                                  & Train              & Test               & Train               & Test              & Train              & Test              \\ \hline
\#Samples                         & 8,855              & 2,933              & 9500                & 2288              & 7200               & 4588              \\ \hline
Caption/image                     & 10                 & 10                 & 10                  & 10                & 10                 & 10                \\ \hline
FID Score                         & \multicolumn{2}{c|}{41.08}              & \multicolumn{2}{c|}{52.34}              & \multicolumn{2}{c|}{58.27}             \\ \hline
\end{tabular}
\end{table}

\subsection{Preprocessing}
For the implementation, we have used Natural Language Toolkit (NLTK) based on the Python platform for statistical natural language processing (NLP). To translate the text from English to Bengali, we used Google Translator.

\subsection{Google translator}
Google Translator is google provided API which can conveniently translate between so far 108 languages by typing. It is fast, free, and has an amazing interface that can recognize almost all existing languages. We use this to translate our input text collected from the CUB dataset. Though it does not provide a fully accurate translation, it was successful for almost 95\% time.

\subsection{Evaluation Metric}
The evaluation of generative models has gotten a lot of attention recently, and numerous quantitative and qualitative methods have been presented so far. The performance of GANs needs to be closely monitored. A few metrics that can be used to verify GANs are as follows:

\subsubsection{\textbf{Quantitative Evaluation}}
\hfill

% \begin{figure*}[h]
% \centering
% \includegraphics[height=200pt, width=480pt]{images/syn-real-imgup.jpg}
% \caption{Comparison between our model generated images and real images}
% \label{syn-real}
% \end{figure*}

\textbf{Inception Score:} The Inception Score \cite{salimans2016improved} is a metric for evaluating the quality of produced images, especially synthetic images generated by generative adversarial network models. It was created in order to reduce subjective human image judgment. The inception score uses a pre-trained deep learning neural network model for image classification to classify the generated images.

\begin{table*}[]
\centering
\caption{Inception scores by other GAN models  
 \cite{reed2016generative,  zhang2017stackgan, zhang2018stackgan++ , reed2016learning} and our Bangla AttnGAN on the CUB test set.}
\label{comparison table}
\begin{tabular}{|c|c|c|c|c|l|}
\hline
\textbf{Dataset} & \textbf{GAN-INT-CLS \cite{reed2016generative}} & \textbf{StackGAN \cite{zhang2017stackgan}} & \textbf{StackGAN-v2 \cite{zhang2018stackgan++}} & \textbf{GAWWN \cite{reed2016learning}} & \multicolumn{1}{c|}{\textbf{Bangla AttnGAN}} \\ \hline
\textbf{CUB}     & 2.88 ± .04                                                      & 3.70 ± .04                                                  & 3.82 ± .06                                                       & 3.62 ± .07                                              & \textbf{3.58 ± .06}                       \\ \hline
\end{tabular}
\end{table*}

\begin{table*}[]
\centering
\caption{Performance of FID and Inception scores by other GAN models \cite{reed2016generative, nguyen2017plug, zhang2017stackgan, zhang2018stackgan++, reed2016learning} and our Bangla-AttnGAN on CUB datasets with Bangla text data.}
\label{table 4}
\begin{tabular}{|c|c|c|c|c|c|}
\hline
\multirow{2}{*}{\textbf{Medium}} & \multirow{2}{*}{\textbf{References}}             & \multirow{2}{*}{\textbf{Dataset}} & \multirow{2}{*}{\textbf{Technique}} & \multicolumn{2}{c|}{\textbf{Measurement}} \\ \cline{5-6} 
                                 &                                             &                                   &                                     & \textbf{FID}   & \textbf{Inception Score} \\ \hline
\multirow{5}{*}{Bangla}          & \cite{reed2016generative}  & CUB                               & GAN-INT-CLS                         & 134.23         & 2.01 ± .02               \\ \cline{2-6} 
                                 & \cite{zhang2017stackgan}   & CUB                               & StackGAN                            & 73.02          & 2.94 ± .09               \\ \cline{2-6} 
                                 & \cite{zhang2018stackgan++} & CUB                               & StackGAN-v2                         & 58.33          & 3.16 ± .02               \\ \cline{2-6} 
                                 & \cite{reed2016learning}    & CUB                               & GAWWN                               & 98.91          & 2.59 ± .06               \\ \cline{2-6} 
                                 &                                             & \textbf{CUB}                      & \textbf{Bangla AttnGAN}             & \textbf{41.08} & \textbf{3.58 ± .06}      \\ \hline
\end{tabular}
\end{table*}

\textbf{FID Score:} Because of its concordance with human inspection and sensitivity to tiny changes in the real distribution, which is based on the Inception v3 Network, Frechet Inception Distance (FID) \cite{heusel2017gans} has achieved significant usage. It is used to ensure that the images created are of high quality and uniformity. As the FID drops, the quality improves. To put it another way, authentic and produced images are strikingly similar. FID score can be calculated using:

\begin{equation}
    {\displaystyle {\text{FID}}=|\mu -\mu _{w}|^{2}+\operatorname {tr} (\Sigma +\Sigma _{w}-2(\Sigma \Sigma _{w})^{1/2}).}
\end{equation}

Squaring Wasserstein metric between two multidimensional Gaussian distributions: $
    {\displaystyle {\mathcal {N}}(\mu ,\Sigma )}{\displaystyle {\mathcal {N}}(\mu ,\Sigma )}
$, FID score is calculated. The distribution of some neural network elements of the pictures produced by the GAN and $
    {\displaystyle {\mathcal {N}}(\mu _{w},\Sigma _{w})}{\displaystyle {\mathcal {N}}(\mu _{w},\Sigma _{w})}
$
And the distribution of similar neural network highlights from the real-world images that were used to create the GAN When the produced and real photos are given as input to the Inception network, it is measured from the mean and covariance of the initiations.

% \textbf{R-Precision:}: Following Xu et al. \cite{xu2018attngan}, we have used R-precision to see if a generated image is well conditioned by the text description. The R-precision is calculated by finding appropriate text in response to an image query. The more visual-semantic similarity between generated images and input text is reflected in a higher score.

\subsubsection{\textbf{Qualitative evaluation}}
\hfill

\begin{figure}[h]
\centering
\includegraphics[height=200pt, width=255pt]{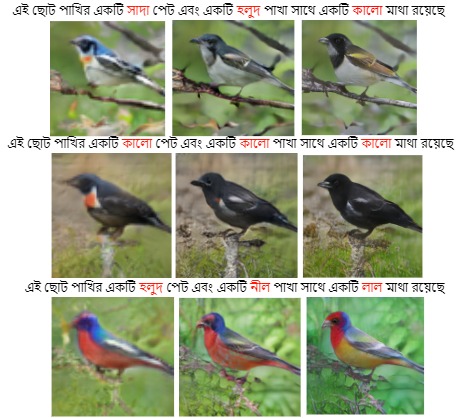}
\caption{Experiment results of our Bangla AttnGAN model trained on CUB dataset with some of the most frequently used terms in the sentence changed}
\label{changing some of the most attended words}
\end{figure}

\textbf{Generalization Ability:}  
Through our experiments, we have demonstrated the AttnGAN's ability to generalize by creating images from text descriptions. 
Fig. \ref{syn-real} shows a few examples. The difference between real and model-generated images can be seen here. It illustrates that these images are modified following our text description. On the other hand, we also notice that images generated by AttnGAN are sharp with great detail and very close to realism.
Moreover, as shown in Fig. \ref{changing some of the most attended words}, after modifying some of the most frequently used words in the text descriptions to see how sensitive the results are to reforms in the inputs, it demonstrates that our proposed model is capable of detecting subtle semantic variations in text descriptions.
We see the AttnGAN's intermediary outcomes with attention to better grasp what the AttnGAN has learned. As shown in Fig. \ref{sada-kalo-img},
The AttnGAN's initial stage just paints the basic size, structure, and tinctures of objects and provides poor-detailed images. After that, the next stages learn how to fix flaws in previous stage outcomes and add more information to create fine-detailed photos.

To summarize, our proposed Bangla AttnGAN's generalization ability is demonstrated by the findings provided in Fig. \ref{syn-real}, \ref{changing some of the most attended words} and \ref{sada-kalo-img}.

% \begin{figure*}[h]
% \centering
% \includegraphics[height=250pt, width=400pt]{images/syn-real-imgg.jpg}
% \caption{Comparison between our model generated images and real images}
% \label{syn-real}
% \end{figure*}

% \begin{figure*}[h]
% \centering
% \includegraphics{images/syn-real-imgup.jpg}
% \caption{Comparison between our model generated images and real images}
% \label{syn-real}
% \end{figure*}

% eta updated pic

% \begin{figure}[h]
% \centering
% \includegraphics[height=200pt, width=255pt]{images/changed text up.jpg}
% \caption{Example results of our AttnGAN model trained on CUB while changing some of the most attended words in the text descriptions}
% \label{changing some of the most attended words}
% \end{figure}

% \begin{figure}[h]
% \centering
% \includegraphics[height=200pt, width=255pt]{images/sada-kalo-img.jpg}
% \caption{More examples of our AttnGAN on CUB as a supplemental material. The first line gives the low-to-high resolution pictures; the second and third line shows the top-5 most-attended words.}
% \label{sada-kalo-img}
% \end{figure}

% \subsubsection{\textbf{Ablation Study}} ******

\subsubsection{\textbf{Comparison with previous methods}}
%  Text-to-image generation on the CUB dataset is compared between AttnGAN and previous state-of-the-art GAN models. On the basis of the CUB dataset, as shown in Table \ref{comparison table}, compared to the previous best score of "****-3.82", "***-our AttnGAN" achieves a "***-4.36" inception score. On the CUB dataset, % Figures 4 and 6 show that 
% our AttnGAN is able to generate images of 256x256 pixels for various scenarios. . It has been shown in the experiments that our AttnGAN model produces more complex images of higher quality than other approaches for its novel attention mechanism.

Text-to-image generation on the  CUB  dataset with Bangla text description is compared between Bangla AttnGAN and previous state-of-the-art  GAN  models.  Based on the  CUB  dataset,  as shown in  Table \ref{dataset table} and \ref{comparison table}, our  Bangla AttnGAN achieves a 3.58 ± .06 Inception score and  41.08 FID score. On the other hand, Table \ref{table 4} demonstrates that FID and Inception score performance on CUB datasets containing Bangla text data using state-of-the-art GAN models \cite{reed2016generative, nguyen2017plug, zhang2017stackgan, zhang2018stackgan++, reed2016learning} and our Bangla-AttnGAN. On the  CUB  dataset, our Bangla AttnGAN can generate images of  256x256  pixels for various scenarios. It has been shown in the experiments that our Bangla AttnGAN model produces more complex images of higher quality than other approaches for its novel attention mechanism.

% \usepackage{multirow}

\section{Conclusion and Future Work}
We present Bangla AttnGAN for high-resolution Bangla text-to-image analysis and generation in this study. To generate high-quality photos, we first develop a one-of-a-kind attentional generating network. Image quality is ensured using a multi-stage procedure. Second, a deep attentional multi-modal similarity model is presented for training the generator, which can compute the generated exquisite image-text matching loss. We outperformed the prior GAN model for text-to-image synthesis by 3.58 ± .06, outperforming the best-announced inception score. Moreover, this is one of the first works collaborating the GAN model with the Bangla language. Our large-scale experimental analysis effectively exemplifies the potency and efficacy of our proposed model. 
We have a plan to work on a more complex COCO dataset as no such work has been processed in this area on the Bangla language. Moreover, we will look forward to analyzing more complex structure text in Bangla.
% \newpage

% \newpage

\bibliographystyle{IEEEtran}
\bibliography{bibliography}

\begin{thebibliography}{10}
\providecommand{\url}[1]{#1}
\csname url@samestyle\endcsname
\providecommand{\newblock}{\relax}
\providecommand{\bibinfo}[2]{#2}
\providecommand{\BIBentrySTDinterwordspacing}{\spaceskip=0pt\relax}
\providecommand{\BIBentryALTinterwordstretchfactor}{4}
\providecommand{\BIBentryALTinterwordspacing}{\spaceskip=\fontdimen2\font plus
\BIBentryALTinterwordstretchfactor\fontdimen3\font minus
  \fontdimen4\font\relax}
\providecommand{\BIBforeignlanguage}[2]{{%
\expandafter\ifx\csname l@#1\endcsname\relax
\typeout{** WARNING: IEEEtran.bst: No hyphenation pattern has been}%
\typeout{** loaded for the language `#1'. Using the pattern for}%
\typeout{** the default language instead.}%
\else
\language=\csname l@#1\endcsname
\fi
#2}}
\providecommand{\BIBdecl}{\relax}
\BIBdecl

\bibitem{goodfellow2014generative}
I.~Goodfellow, J.~Pouget-Abadie, M.~Mirza, B.~Xu, D.~Warde-Farley, S.~Ozair,
  A.~Courville, and Y.~Bengio, ``Generative adversarial nets,'' \emph{Advances
  in neural information processing systems}, vol.~27, 2014.

\bibitem{isola2017image}
P.~Isola, J.-Y. Zhu, T.~Zhou, and A.~A. Efros, ``Image-to-image translation
  with conditional adversarial networks,'' in \emph{Proceedings of the IEEE
  conference on computer vision and pattern recognition}, 2017, pp. 1125--1134.

\bibitem{zhu2017unpaired}
J.-Y. Zhu, T.~Park, P.~Isola, and A.~A. Efros, ``Unpaired image-to-image
  translation using cycle-consistent adversarial networks,'' in
  \emph{Proceedings of the IEEE international conference on computer vision},
  2017, pp. 2223--2232.

\bibitem{yi2017dualgan}
Z.~Yi, H.~Zhang, P.~Tan, and M.~Gong, ``Dualgan: Unsupervised dual learning for
  image-to-image translation,'' in \emph{Proceedings of the IEEE international
  conference on computer vision}, 2017, pp. 2849--2857.

\bibitem{reed2016generative}
S.~Reed, Z.~Akata, X.~Yan, L.~Logeswaran, B.~Schiele, and H.~Lee, ``Generative
  adversarial text to image synthesis,'' in \emph{International Conference on
  Machine Learning}.\hskip 1em plus 0.5em minus 0.4em\relax PMLR, 2016, pp.
  1060--1069.

\bibitem{zhang2017stackgan}
H.~Zhang, T.~Xu, H.~Li, S.~Zhang, X.~Wang, X.~Huang, and D.~N. Metaxas,
  ``Stackgan: Text to photo-realistic image synthesis with stacked generative
  adversarial networks,'' in \emph{Proceedings of the IEEE international
  conference on computer vision}, 2017, pp. 5907--5915.

\bibitem{xu2018attngan}
T.~Xu, P.~Zhang, Q.~Huang, H.~Zhang, Z.~Gan, X.~Huang, and X.~He, ``Attngan:
  Fine-grained text to image generation with attentional generative adversarial
  networks,'' in \emph{Proceedings of the IEEE conference on computer vision
  and pattern recognition}, 2018, pp. 1316--1324.

\bibitem{zhang2018photographic}
Z.~Zhang, Y.~Xie, and L.~Yang, ``Photographic text-to-image synthesis with a
  hierarchically-nested adversarial network,'' in \emph{Proceedings of the IEEE
  Conference on Computer Vision and Pattern Recognition}, 2018, pp. 6199--6208.

\bibitem{miyato2018cgans}
T.~Miyato and M.~Koyama, ``cgans with projection discriminator,'' \emph{arXiv
  preprint arXiv:1802.05637}, 2018.

\bibitem{odena2017conditional}
A.~Odena, C.~Olah, and J.~Shlens, ``Conditional image synthesis with auxiliary
  classifier gans,'' in \emph{International conference on machine
  learning}.\hskip 1em plus 0.5em minus 0.4em\relax PMLR, 2017, pp. 2642--2651.

\bibitem{zhang2018stackgan++}
H.~Zhang, T.~Xu, H.~Li, S.~Zhang, X.~Wang, X.~Huang, and D.~N. Metaxas,
  ``Stackgan++: Realistic image synthesis with stacked generative adversarial
  networks,'' \emph{IEEE transactions on pattern analysis and machine
  intelligence}, vol.~41, no.~8, pp. 1947--1962, 2018.

\bibitem{wah2011caltech}
C.~Wah, S.~Branson, P.~Welinder, P.~Perona, and S.~Belongie, ``The caltech-ucsd
  birds-200-2011 dataset,'' 2011.

\bibitem{qiao2019mirrorgan}
T.~Qiao, J.~Zhang, D.~Xu, and D.~Tao, ``Mirrorgan: Learning text-to-image
  generation by redescription,'' in \emph{Proceedings of the IEEE/CVF
  Conference on Computer Vision and Pattern Recognition}, 2019, pp. 1505--1514.

\bibitem{naveen2021transformer}
S.~Naveen, M.~R. Kiran, M.~Indupriya, T.~Manikanta, and P.~Sudeep,
  ``Transformer models for enhancing attngan based text to image generation,''
  \emph{Image and Vision Computing}, p. 104284, 2021.

\bibitem{sharma2018chatpainter}
S.~Sharma, D.~Suhubdy, V.~Michalski, S.~E. Kahou, and Y.~Bengio, ``Chatpainter:
  Improving text to image generation using dialogue,'' \emph{arXiv preprint
  arXiv:1802.08216}, 2018.

\bibitem{li2019controllable}
B.~Li, X.~Qi, T.~Lukasiewicz, and P.~H. Torr, ``Controllable text-to-image
  generation,'' \emph{arXiv preprint arXiv:1909.07083}, 2019.

\bibitem{yin2019semantics}
G.~Yin, B.~Liu, L.~Sheng, N.~Yu, X.~Wang, and J.~Shao, ``Semantics
  disentangling for text-to-image generation,'' in \emph{Proceedings of the
  IEEE/CVF Conference on Computer Vision and Pattern Recognition}, 2019, pp.
  2327--2336.

\bibitem{hu2021crd}
T.~Hu, C.~Long, and C.~Xiao, ``Crd-cgan: Category-consistent and relativistic
  constraints for diverse text-to-image generation,'' \emph{arXiv preprint
  arXiv:2107.13516}, 2021.

\bibitem{ramesh2021zero}
A.~Ramesh, M.~Pavlov, G.~Goh, S.~Gray, C.~Voss, A.~Radford, M.~Chen, and
  I.~Sutskever, ``Zero-shot text-to-image generation,'' \emph{arXiv preprint
  arXiv:2102.12092}, 2021.

\bibitem{li2020manigan}
B.~Li, X.~Qi, T.~Lukasiewicz, and P.~H. Torr, ``Manigan: Text-guided image
  manipulation,'' in \emph{Proceedings of the IEEE/CVF Conference on Computer
  Vision and Pattern Recognition}, 2020, pp. 7880--7889.

\bibitem{xia2021tedigan}
W.~Xia, Y.~Yang, J.-H. Xue, and B.~Wu, ``Tedigan: Text-guided diverse face
  image generation and manipulation,'' in \emph{Proceedings of the IEEE/CVF
  Conference on Computer Vision and Pattern Recognition}, 2021, pp. 2256--2265.

\bibitem{khan2020realistic}
M.~Z. Khan, S.~Jabeen, M.~U.~G. Khan, T.~Saba, A.~Rehmat, A.~Rehman, and
  U.~Tariq, ``A realistic image generation of face from text description using
  the fully trained generative adversarial networks,'' \emph{IEEE Access},
  vol.~9, pp. 1250--1260, 2020.

\bibitem{schulze2021cagan}
H.~Schulze, D.~Yaman, and A.~Waibel, ``Cagan: Text-to-image generation with
  combined attention gans,'' \emph{arXiv preprint arXiv:2104.12663}, 2021.

\bibitem{huang2015bidirectional}
Z.~Huang, W.~Xu, and K.~Yu, ``Bidirectional lstm-crf models for sequence
  tagging,'' \emph{arXiv preprint arXiv:1508.01991}, 2015.

\bibitem{reed2016learning}
S.~E. Reed, Z.~Akata, S.~Mohan, S.~Tenka, B.~Schiele, and H.~Lee, ``Learning
  what and where to draw,'' \emph{Advances in neural information processing
  systems}, vol.~29, pp. 217--225, 2016.

\bibitem{lin2014microsoft}
T.-Y. Lin, M.~Maire, S.~Belongie, J.~Hays, P.~Perona, D.~Ramanan,
  P.~Doll{\'a}r, and C.~L. Zitnick, ``Microsoft coco: Common objects in
  context,'' in \emph{European conference on computer vision}.\hskip 1em plus
  0.5em minus 0.4em\relax Springer, 2014, pp. 740--755.

\bibitem{salimans2016improved}
T.~Salimans, I.~Goodfellow, W.~Zaremba, V.~Cheung, A.~Radford, and X.~Chen,
  ``Improved techniques for training gans,'' \emph{Advances in neural
  information processing systems}, vol.~29, pp. 2234--2242, 2016.

\bibitem{nguyen2017plug}
A.~Nguyen, J.~Clune, Y.~Bengio, A.~Dosovitskiy, and J.~Yosinski, ``Plug \& play
  generative networks: Conditional iterative generation of images in latent
  space,'' in \emph{Proceedings of the IEEE Conference on Computer Vision and
  Pattern Recognition}, 2017, pp. 4467--4477.

\bibitem{heusel2017gans}
M.~Heusel, H.~Ramsauer, T.~Unterthiner, B.~Nessler, and S.~Hochreiter, ``Gans
  trained by a two time-scale update rule converge to a local nash
  equilibrium,'' \emph{Advances in neural information processing systems},
  vol.~30, 2017.

\end{thebibliography}

\end{document}